\documentclass{INTERSPEECH2023}

\usepackage{multirow}
\usepackage{cite}
\usepackage{float}
\floatstyle{plaintop}
\restylefloat{table}
\usepackage[inkscapearea=page]{svg}
\newcommand{\mytextapprox}{\raisebox{0.5ex}{\texttildelow}}
\newcommand{\comment}[1]{}

\interspeechcameraready

\title{Multilingual context-based pronunciation learning for Text-to-Speech}
\name{Giulia Comini, Manuel Sam Ribeiro, Fan Yang, Heereen Shim, Jaime Lorenzo-Trueba}
\address{Amazon Alexa, TTS Research}
\email{
    \{gcomini, manuerib, fffyang, shimreen, truebaj\}@amazon.com}

\begin{document}

\maketitle
 
\begin{abstract}
Phonetic information and linguistic knowledge are an essential component of a Text-to-speech (TTS) front-end. Given a language, a lexicon can be collected offline and Grapheme-to-Phoneme (G2P) relationships are usually modeled in order to predict the pronunciation for out-of-vocabulary (OOV) words. Additionally, post-lexical phonology, often defined in the form of rule-based systems, is used to correct pronunciation within or between words. 
In this work we showcase a multilingual unified front-end system that addresses any pronunciation related task, typically handled by separate modules. We evaluate the proposed model on G2P conversion and other language-specific challenges, such as homograph and polyphones disambiguation, post-lexical rules and implicit diacritization. We find that the multilingual model is competitive across languages and tasks, however, some trade-offs exists when compared to equivalent monolingual solutions.

\end{abstract}

\section{Introduction}
Traditionally, the front-end of a Text-to-Speech (TTS) system is a cascade of modules that cover specific linguistic tasks. These may include Grapheme-to-Phoneme (G2P) conversion, post-lexical phonology, text normalization, among others. Each of these tasks can either be a collection of hand-crafted rules, (weighted) finite-state transducer solutions, or, eventually, statistical models. This long-established solution addresses tasks independently, introducing dependencies between components that can be detrimental to efficient maintenance. 

Recent studies have been focusing on improving one or more front-end components and the main areas of interest are: 1) approaches to address specific tasks (e.g. \cite{bisaniN08, novakMH12, peters17, multimodal19, amazon20, byt522, Yolchuyeva_2019, char_transformer20, gbert22} for G2P conversion); 
2) the realisation of a unified front-end \cite{gorman-etal-2018-improving, T5G2P_21, apple_frontend20, soundchoice22}; 3) the use of multilingual knowledge to improve performances \cite{deriK16, peters17, multimodal19, amazon20, apple_frontend20, byt522}; 4) zero/low-resource language expansion \cite{byt522, gbert22, sigmorphon20, zeroshot22}. 
Examples of approaches that combine more than one front-end task are \cite{apple_frontend20}, which proposes a multilingual neural system for both G2P and text normalization, and \cite{soundchoice22}, which tweaks a neural model trained on the G2P task to learn to effectively perform homograph disambiguation on English data.

Inspired by these efforts, we propose a \emph{multilingual context-based pronunciation model for Text-to-Speech systems}.
This is a unified data-driven solution that addresses any pronunciation related task traditionally handled by independent TTS front-end modules. 

Given the promising results in multilingual G2P literature \cite{peters17, multimodal19, amazon20, apple_frontend20, byt522, zeroshot22}, we use multilingual pronunciation knowledge
to train our system. We leverage pronunciation data not only at word level, like it is traditionally done for G2P conversion \cite{peters17, Yolchuyeva_2019, char_transformer20, amazon20, gbert22, byt522}, but also at sentence level, similarly to \cite{gorman-etal-2018-improving, multimodal19, apple_frontend20, T5G2P_21, soundchoice22}. This allows us to go beyond G2P conversion and set a step towards a fully learnable neural front-end.

We evaluate our model on the G2P task for 24 languages and dialects, at the word and sentence level, comparing against equivalent monolingual baselines.
Additionally, we evaluate the system on language-specific pronunciation tasks, such as homograph disambiguation for British and American English, polyphone disambiguation for Mandarin, post-lexical rules for European French, and implicit diacritization for Standard and Saudi Arabian Arabic. 
We find that the multilingual model achieves competitive results across all tasks. However, we highlight that there exists trade-offs between multilingual and monolingual solutions, likely depending on the task, the language and, most importantly, the data quality and availability.

The paper is structured as follows: Section \ref{sec:g2p} describes the data, their format, the evaluation methodology and the proposed model architecture; Section \ref{sec:exp} presents the results on the G2P task across languages and analyses the above mentioned language-specific challenges. It additionally shows the impact of using either multilingual or monolingual data for all the considered tasks; Section \ref{sec:conclusions} concludes.

\begin{figure*}[t]
  \centering
  \includegraphics[width=1.0\linewidth]{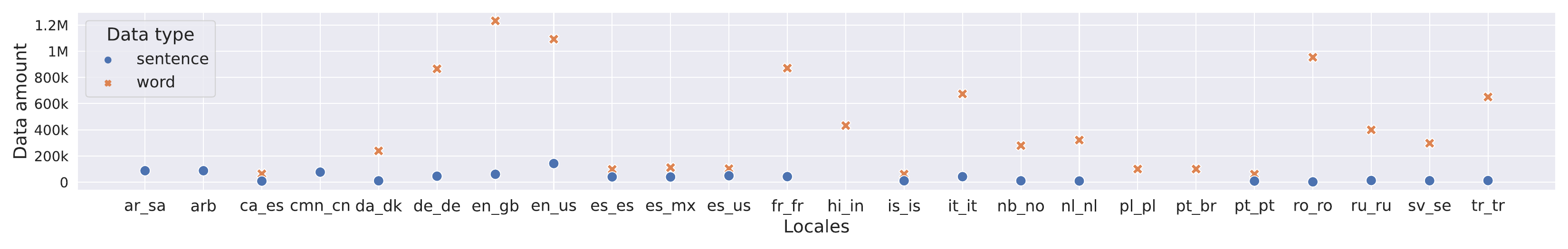}
  \caption{Sentence and word level amount of data. Note that for some locales we only have data for one of the two data types.}
  \label{fig:amount}
\end{figure*}

\begin{figure*}[t]
  \centering
  \includegraphics[width=1.0\linewidth]{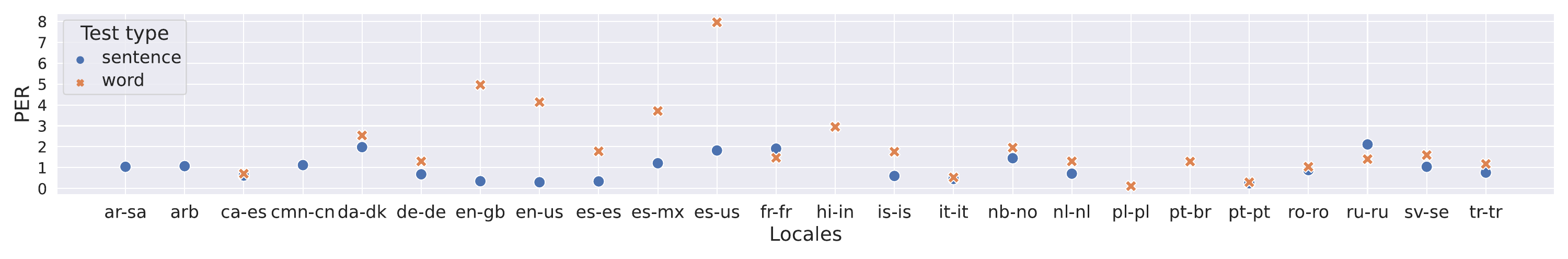}
  \caption{Phone Error Rate (PER\%) for the multilingual G2P system at the sentence and word levels.}
  \label{fig:all}
\end{figure*}

\section{Multilingual pronunciation learning}
\label{sec:g2p}

\textbf{Pronunciation data.}
We collect pronunciation data for 24 unique locales. The data consist of \textit{$<$text, pronunciation$>$} pairs, where \textit{text} can be either a single word or a sentence. 
The word level data consist of internal pronunciation dictionaries (lexica) across 21 locales. On average, they contain \mytextapprox314k word pronunciations,
ranging from 50k up to 1.2M entries. The lexica are a collection of multiple sources and they include person names, place names, foreign words, and other domain-specifc entries. We additionally generate pronunciation at the sentence level for 18 locales, using internally available front-ends. For another 3 locales, Standard Arabic (\texttt{arb}), Saudi Arabian Arabic (\texttt{ar-sa}) and Mandarin (\texttt{cnm-cn}), we use human-annotated sentence level data, but we do not have lexica. 
Figure \ref{fig:amount} shows the amount of word and sentence level data per locale. 
We use the X-SAMPA phonetic transcription scheme to represent phonetic information. We include X-SAMPA stress and diacritics markers, but exclude syllable boundaries.

\textbf{Test sets creation.}
Because our system operates at the word and sentence level, we construct two types of test sets.
For the test set at the word level, we apply an 85/5/10 split for train, development and test sets using individual words from the lexica.
This test set is created independently for each locale, following two criteria: 1) all the words with the same lemma belong to the same train/development/test partition\footnote{We opt for this partition following the evaluation design described in the SIGMORPHON 2020 shared task on multilingual G2P conversion \cite{sigmorphon20}}; 2) most of the words in the test set should not be common words in that locale. 

Our aim is to have fair evaluation mechanism and to simulate a real-world scenario in which only the most common words are annotated. We rely on the Spacy library \cite{spacy2} to extract word lemmas for the supported languages and on heuristics for the non-supported ones. We get words frequencies information from the multilingual C4 corpus\footnote{https://www.tensorflow.org/datasets/catalog/c4\#c4multilingual} (internally cleaned). We randomly sample the words frequency distribution, limited up to the 95th percentile, and add to the test set all the words with the same lemma of the sampled word. The test set might still contain relatively common words, if they share a lemma with the selected uncommon words. The same procedure is applied for the generation of the development set. On average, the word level test set contains \mytextapprox30k words per locale. 

It is less trivial to design a fair evaluation set at the sentence level.
For simplicity, the sentence level test set consists of a random selection of 1-10\% of the per-locale sentence data.

\textbf{Input/output data format.}
In order to handle multilingual information, we prepend languages codes to the input entries, as recommended in \cite{langid22}. Every locale has a unique code, defined by the W3C language identification tags (ISO 639-1 for the language name and ISO 3166-1 alpha 2 for the country code). The input text is provided at character level, therefore tokenization is not necessary. This allows us to keep a relatively small vocabulary size. For the sentence data, in the absence of punctuation, a word-boundary token is used to represent the space in between words. The target sequences consist of space-separated phonemes in the X-SAMPA format, with the word-boundary token in between words in absence of punctuation. Syllable breaks are discarded. The output vocabulary includes phonemes with X-SAMPA symbols for primary or secondary stress markers, or diacritics, as unique entries.

\textbf{Model specifications.}
We use a 12-layer transformer encoder-decoder model, as described in \cite{vaswani17}. Transformers have been proved successful in sequence-to-sequence tasks, including G2P conversion \cite{Yolchuyeva_2019, multimodal19, amazon20, char_transformer20, apple_frontend20, gbert22}. Leveraging the OpenNMT \cite{opennmt} implementation of transformer, we train the model for 1M steps on 6 GPUs (Tesla V100-SXM2-32GB), with dynamic batching for up to 4096 tokens. As in \cite{vaswani17}, we use \textit{Adam} as optimizer and ``noam" as learning rate schedule, with \textit{warmup\_steps}=8000. Our model contains \mytextapprox52.5M parameters.

\begin{table*}[t]
\centering
\begin{tabular}{ccccc}
\hline
 &

\multicolumn{2}{c}{\textbf{Monolingual}} &
\multicolumn{2}{c}{\textbf{Multilingual}} \\

&
\textbf{Words (PER/WER)} &
\textbf{Sentences (PER/SER)} &
\textbf{Words (PER/WER)} &
\textbf{Sentences (PER/SER)} \\

\hline
cmn-cn & 
/ & 
\textbf{1.07/31.07} & 
/ & 
1.12/35.47 \\

en-gb & 
7.86/27.76 & 
0.74/18.87 & 
\textbf{4.96/18.57} & 
\textbf{0.35/10.47} \\

en-us & 
6.74/24.16 & 
0.35/9.02 & 
\textbf{4.14/16.18} & 
\textbf{0.30/7.97}\\

fr-fr & 
3.08/10.01 & 
\textbf{1.76/40.28} & 
\textbf{1.47/5.71} & 
1.88/44.04  \\
\hline
\end{tabular}
\caption{Monolingual VS multilingual models' phone, word or sentence error rates (\%) on OOV words and sentences on 4 locales.}
\label{table:results}
\end{table*}

\section{Experiments}
\label{sec:exp}

We train a multilingual G2P model on pronunciation data for 24 locales. We choose the best evaluation step looking at the average performance on the development set across locales. We evaluate the model using the following metrics: word-error-rate (WER\%) on the word level test set, sentence-error-rate (SER\%) on the sentence level test set and phone-error-rate (PER\%) on both test sets. Additionally, we analyse language-specific challenges, such as homograph disambiguation for British and American English (\texttt{en-gb}, \texttt{en-us}), polyphone disambiguation for Mandarin (\texttt{cmn-cn}), post-lexical rules for French (\texttt{fr-fr}) and diacrization for the Arabic locales (\texttt{arb} and \texttt{ar-sa}). For these locales we also train equivalent monolingual models, in order to understand the impact of the multilingual knowledge on every task.

\subsection{Model performances on the G2P task}
\label{subsec:results}
Figure \ref{fig:all} shows the PERs on the sentence and word level test sets across all locales. On average across locales, the PER on the word level test set is 2.09\%, while on the sentence level test set is 0.97\%. Higher performances on sentences are expected, since not all words in this test set are out-of-vocabulary (OOV).
On average, the WER on the word level test set is 8.45\%, while the SER on the sentence level test set is 27.55\%.
At word level, we obtain the highest PERs on \texttt{en-gb}, \texttt{en-us}, \texttt{es-mx} and \texttt{es-us}. The English and Spanish lexica contain foreign and rare domain-specific words, which are challenging and tend to negatively affect the global performance. Overall, we believe that the results on OOV words are competitive, considering that the test set is composed by words with unseen lemmas and that we also generate X-SAMPA stress markers and diacritics.

We compare the proposed multilingual system with equivalent monolingual models for four locales: \texttt{cmn-cn}, \texttt{en-gb}, \texttt{en-us} and \texttt{fr-fr}.
Table \ref{table:results} shows that a multilingual model tends to outperform a monolingual one.
We observe improvements for \texttt{en-gb}, \texttt{en-us}, and \texttt{fr-fr} on the word level test set.
For Mandarin (\texttt{cmn-cn}), the monolingual model outperforms the multilingual one. We hypothesise that Mandarin might not benefit from the other locales knowledge, since it has a different writing system, it doesn't have word boundaries, it is a tone language and it is highly homographic.

Regarding French, sentence level data poses the additional challenge of post-lexical rules, a phenomenon which we will discuss in Section \ref{subsubsec:french}.

\subsection{Language-specific analysis}
\label{subsec:analysis}
We propose a unified front-end system that is capable of addressing all pronunciation-related tasks, typically handled by separate modules. We analyse the multilingual model performances across a set of languages and tasks, and compare them to equivalent monolingual models. 
Specifically, we consider homograph disambiguation for English, polyphone disambiguation for Mandarin, post-lexical rules for French, and implicit diacritization for Arabic.
We choose these tasks because they are context-dependent and address phenomena that is not trivially handled by word level pronunciation dictionaries.

\begin{table}[t]
\centering
\begin{tabular}{ccccc}

\hline
 &

\multicolumn{2}{c}{\textbf{Homograph dis. accuracy (\%)}} \\
 &
\textbf{Monolingual} &
\textbf{Multilingual}

&
\textbf{\# words}
\\

\hline

en-gb &  
95.49 &
\textbf{96.59} &
821
\\

en-us & 
94.5 &
\textbf{95.88} &
800
\\
\hline

\end{tabular}
\caption{Homograph disambiguation  accuracy (\%) for English locales. Results are shown for both the monolingual and multilingual models, at their respective best steps, on a common subset of the homographs evaluation set.}
\label{table:english}
\end{table}

\subsubsection{Homograph disambiguation for English}
\label{subsubsec:english}
English homograph disambiguation has more often been approached in literature as a classification problem \cite{gorman-etal-2018-improving, nicolis21_ssw}, with recent work exploring it jointly with the G2P task \cite{T5G2P_21, soundchoice22}. The latter studies use neural monolingual G2Ps operating at the sentence level: \cite{T5G2P_21} uses a vanilla transformer architecture, but only analyses the accuracy on 3 English homographs, while \cite{soundchoice22} uses curriculum learning and an homograph disambiguation loss to improve the disambiguation task on top of the G2P task.

We aim to have a fair representation of homographs during training, following \cite{gorman-etal-2018-improving, nicolis21_ssw, soundchoice22}.
We therefore make use of the open source homograph data released by \cite{gorman-etal-2018-improving}\footnote{https://github.com/google/WikipediaHomographData}, consisting of \mytextapprox14.5k sentences for training and \mytextapprox1.6k sentences for evaluation.
We generate pronunciations at the sentence level for both \texttt{en-gb} and \texttt{en-us} using our internal front-ends.
We focus the evaluation on the homograph disambiguation accuracy, comparing the homographs' gold and predicted X-SAMPA pronunciations. Unfortunately in a number of cases we aren't able to accurately isolate the phoneme sequences of the homographs in the gold and predicted sequences, therefore we evaluate only on a common subset of the available test data (see Table \ref{table:english}). Table \ref{table:english} shows that the multilingual model outperforms the respective monolingual model in the homograph disambiguation task for both locales. This is in line with the results of Section \ref{subsec:results} on the G2P task.

\begin{table}[t]
\centering
\begin{tabular}{cccc}

\hline
 &

\multicolumn{2}{c}{\textbf{Character pronunciation accuracy (\%)}} \\
 &
\textbf{Whole sentence} &
\textbf{Polyphones only}
\\

\hline

Monolingual &  
\textbf{98.80} &
\textbf{98.67}
\\

Multilingual & 
98.15 &
97.24
\\
\hline

\end{tabular}
\caption{Character pronunciation accuracy (\%) for Mandarin. Results are shown for both the monolingual and multilingual models, on the entire sentence and on the polyphonic characters only.}%
\label{table:mandarin}
\end{table}

\begin{table}[t]
\centering
\begin{tabular}{cccc}

\hline
 &

\multicolumn{2}{c}{\textbf{PER/WER}} \\
 &
\textbf{Whole sentence} &
\textbf{PLRs only}
\\

\hline

Monolingual &  
\textbf{1.76}/ - &
\textbf{1.67/6.26}
\\

Multilingual & 
1.88/ - &
3.14/8.54
\\
\hline

\end{tabular}
\caption{Phone and word error rates (\%) on the French sentence level test set. Results are shown for both the monolingual and multilingual models, on the entire sentence and on the words affected by post-lexical rules.}
\label{table:french}
\end{table}

\begin{table*}[t]
\centering
\begin{tabular}{ccccc}
\hline
 &

\multicolumn{2}{c}{\textbf{Monolingual (PER/SER)}} &
\multicolumn{2}{c}{\textbf{Multilingual (PER/SER)}} \\

&
\textbf{with diacritics} &
\textbf{w/o diacritics} &
\textbf{with diacritics} &
\textbf{w/o diacritics} \\

\hline
arb & 
0.46/13.00 & 
\textbf{1.73/50.00} & 
\textbf{0.29/9.00} & 
1.85/54.50 \\

ar-sa & 
0.57/18.50 & 
1.82/\textbf{51.00} & 
\textbf{0.32/8.50} & 
\textbf{1.76}/51.50 \\

\hline
\end{tabular}
\caption{Monolingual VS multilingual models' phone and sentence error rates (\%) for Arabic locales, when the input script is diacritized and undiacritized.}
\label{table:arabic}
\end{table*}

\subsubsection{Polyphone disambiguation for Mandarin}
\label{subsubsec:mandarin}
Mandarin (\texttt{cmn-cn}) writing system uses a morphosyllabic script, where some characters have a 1:1 relationship with phones, while others, called polyphonic characters, can have multiple pronunciations according to their context. 
One of the main challenges for Mandarin G2P conversion is polyphone disambiguation (PD).
Recent solutions that address the PD task in Mandarin include a unified monolingual neural front-end for joint segmentation, text normalization, part-of-speech prediction, and G2P conversion \cite{unified_frontend20}; a pre-trained Chinese BERT-like \cite{bert} language model for polyphone disambiguation \cite{unified_frontend_bert20, zhang21, zhang22, g2pW_chen22}, eventually extended with multi-task training \cite{unified_frontend_bert20, g2pW_chen22}.

In this section we show that our system learns PD without employing any additional technique.
We extract a list of polyphonic characters from the human-annotated training data. We use \mytextapprox71.9k sentences for training, and 750 for evaluation. Note that we don't have a Mandarin lexicon for these experiments. In the test set, out of 15837 characters, 4862 are polyphones (30.7\%), with a coverage of 315 unique polyphonic characters. 

In Table \ref{table:mandarin} we compute the character pronunciation accuracy (\%) for both the monolingual and multilingual models, and compare it with the PD accuracy (\%). Similarly to what we have observed in Section \ref{subsec:results}, the monolingual model outperforms the multilingual one also in the PD task.
Our results cannot be directly compared to others in the recent literature due to differences in the training and test data. However, we do highlight that the PD accuracy is aligned with recent work using more sophisticated neural solutions, often relying on amounts of data from 500k+ \cite{unified_frontend_bert20} to 1M+ \cite{unified_frontend20, zhang22, g2pW_chen22} training examples.

\subsubsection{Post-lexical rules for French}
\label{subsubsec:french}
French pronunciation relies on complex relationships between words, and phenomena like enchaînement and liaisons are very common. Enchaînement (or linking) consists in taking the last consonant of a word and pronouncing it with the first syllable of the next word. Liaisons determine the pronunciation of latent consonants, e.g. consonants which aren't pronounced if a word is isolated, but can be pronounced in specific phonetic and syntactic contexts.
Taylor et al. \cite{liaisons21} confirm the importance of post-lexical rules (PLRs) in French TTS, showing that a French TTS voice is preferred when the front-end uses a PLRs module to correct the pronunciations of words, given a context. 

Table \ref{table:results} compares the performances of monolingual and multilingual models on French OOV words and sentences, but those numbers do not give an indication on how the models perform on PLRs. 
We therefore evaluate our unified multilingual front-end for post-lexical phenomena on French sentences.
We construct a gold set using a set of internal PLRs, derived by a French language expert.
Pronunciations are generated at the sentence level, but, for testing purposes, we consider only words which have been modified by PLRs.
We evaluate performance in terms of PERs and WERs.
Note that both the monolingual and multilingual models use \mytextapprox33.7k French sentences, while the rest of the French training set consists of \mytextapprox800k individual words.
The test set contains 4211 sentences, 83.16\% of which are affected by at least one PLR.
Out of 46540 words in the sentence test set, 9877 (21.1\%) are affected by the PLRs module. 

We previously observed that, for French, a monolingual model outperforms a multilingual one on sentence level inputs (Table \ref{table:results}).
The results for PLRs presented in Table \ref{table:french} show the same trend: the monolingual model obtains 1.67\% PER on words affected by PLRs, while the multilingual is at 3.14\% PER.
These results could be partially explained by the limited amount of entries affected by PLRs in the training data for the multilingual scenario.
The multilingual model is likely less capable to learn the nuances which characterise French PLRs.
Balancing sentence level and word level data might help alleviate this problem.
Additionally, we highlight that we do not have data tailored for the PLRs task and it is possible that the random selection of training sentence might have missed specific contexts.
In future work we could design an ad-hoc PLRs set in order to evaluate the model's ability to generalise, given what it has observed during training.

\subsubsection{Diacritization for Arabic locales}
\label{subsubsec:arabic}
When Arabic text is diacritized, the G2P task is mostly trivial. However, Arabic text data is often available in the undiacritized form and a diacritizer becomes a necessary component of the Arabic front-end.
Some recent efforts that build Arabic diacritizers rely on the Viterbi algorithm \cite{Darwish17, hadjir19} or neural network systems \cite{dnn19, s2s19, s2s2_19}.
More recently, \cite{dia_and_translate21} showed that a transformer model \cite{vaswani17} achieves higher performances in the diacritazion task when it's trained in a multi-task fashion to both diacritize and translate.
Additionally, \cite{bert_diacritizer22} proposes to diacritize text using a BERT-like tagger.

In this work, we aim to learn Arabic pronunciation from both diacritized and undiacritized text for two Arabic locales. When the model is given undiacritized input, it must learn the diacritization task.
We use the same \mytextapprox82.2k training sentences for both locales, half of which is undiacritized.
We evaluate on a test set of 400 sentences, with a 50\% split for diacritized and undiacritized sentences. 
Note that for both training and testing the content of the diacritized and undiacritized text is the same.

Table \ref{table:arabic} shows PER and SER for \texttt{arb} and \texttt{ar-sa}, comparing the monolingual and the multilingual systems.
We observe that both locales benefit from multilingual knowledge when the input text is diacritized. However, when the input text is undiacritized, results are mixed. Overall, in terms of PER, the model performs 3-6 times better when the input is diacritized, which is expected. Although more than half of the undiacritized test sentences contains at least an error, we observe that the undiacritized training sentences are of a relatively limited number (\mytextapprox41.1k) and the PERs are still below 2\% for both locales.

\section{Discussion and conclusions}
\label{sec:conclusions}
We present an initial effort towards a fully learnable multilingual neural front-end for TTS. We train a transformer model that generates pronunciations given a word or a normalized text sequence for 24 locales. We demonstrate how the proposed system can effectively predict pronunciation for OOV words with unseen lemmas, as well as learning challenging language-specific tasks such as homograph and polyphone disambiguation, post-lexical phonology and implicit diacritization. 

From our analysis it is evident that multilingual knowledge is beneficial for pronunciation learning, however there are trade-offs with respect to equivalent monolingual solutions. We argue that the model performances on a particular task might be affected by the quality and availability of the relevant data for that task. Nonetheless the results are generally promising and set a step towards a unified neural front-end. To the best of our knowledge, this is the first work which gives such a broad overview over a unified TTS front-end. 

Future work could involve: 1) incorporating text normalization into the picture; 2) understanding which and how much data are necessary to learn a language pronunciation; 3) exploring low-resource language expansion; 4) using the unified neural front-end to build TTS voices across languages; 5) considering memory and latency requirements for real-time applications.

\bibliographystyle{IEEEtran}
\bibliography{mybib}

\end{document}